\documentclass{llncs}
\usepackage{latexsym,amssymb,amsmath,amstext}
\usepackage{graphicx,url,adjustbox,xcolor}
\usepackage{pgfplots}
  \pgfplotsset{width=7cm,compat=1.15}
  \usepgfplotslibrary{statistics}
\usepackage{tikz}
  \usetikzlibrary{positioning,shapes,arrows,fit}
\usepackage{hyperref}

\begin{document}

\title{An Agentic Framework for Rapid Deployment of Edge AI Solutions in Industry 5.0%
  \thanks{Manuscript accepted in PRO-VE 2025; final version available at \url{https://doi.org/10.1007/978-3-032-05681-8_4}.}}
\author{Jorge Martinez-Gil\inst{1} \and Mario Pichler\inst{1} \and Nefeli Bountouni\inst{2} \and Sotiris Koussouris\inst{3} \and Marielena Márquez Barreiro\inst{4} \and Sergio Gusmeroli\inst{5}}
\institute{
  $^{1}$Software Competence Center Hagenberg GmbH, Austria\\
  $^{2}$SUITE5, Gerakas, Greece\\
  $^{3}$SUITE5, Limassol, Cyprus\\
  $^{4}$Gradiant, Vigo, Spain\\
  $^{5}$Politecnico di Milano, Italy\\
  \email{jorge.martinez-gil@scch.at}
}

\maketitle

\begin{abstract}
We present a novel framework for Industry 5.0 that simplifies the deployment of AI models on edge devices in various industrial settings. The design reduces latency and avoids external data transfer by enabling local inference and real-time processing. Our implementation is agent-based, which means that individual agents, whether human, algorithmic, or collaborative, are responsible for well-defined tasks, enabling flexibility and simplifying integration. Moreover, our framework supports modular integration and maintains low resource requirements. Preliminary evaluations concerning the food industry in real scenarios indicate improved deployment time and system adaptability performance. The source code is publicly available at \url{https://github.com/AI-REDGIO-5-0/ci-component}.

\end{abstract}

\section{Introduction}
Industry 5.0 has led to the integration of real-time AI inference at the edge, enabling factory operators to supervise automated systems more closely \cite{hoch2023multi}. This vision builds on Industry 4.0's priority on digital transformation, the Internet of Things (IoT), and Artificial Intelligence (AI). It focuses on embedding smart functions and human involvement in industrial workflows \cite{DBLP:journals/access/BuchgeherGGE21,DBLP:journals/frai/KrausePKKLESUDKDGHHGSM24}. A key element is the convergence of edge computing and AI, commonly known as Edge AI \cite{DBLP:conf/interact/BerghRG23}, which supports real-time data handling at the network edge \cite{DBLP:journals/cm/ShenSZLPLZL24}. This combination is essential for enabling responsive, human-centered manufacturing processes.

Shifting AI tasks from cloud servers to local devices reduces delay, cuts down on network load, and limits the exposure of sensitive data. This change is critical for applications requiring fast responsiveness \cite{DBLP:journals/internet/MeuserLBPDABBLDEGMPRSSS24,DBLP:conf/ictc/SampedroMSA0L22,DBLP:journals/ijon/SunWMZ25}, where prompt actions are essential. As demand for such systems increases, organizations require solutions to speed up Edge AI deployment \cite{DBLP:journals/access/DonnoTD19,DBLP:conf/icccn/VaterHK19}.

Local data processing also improves service quality which is critical for many applications \cite{DBLP:journals/jiii/GauttamPBNP23,DBLP:conf/case/RupprechtHV22}. There is a broad consensus that Edge AI increases efficiency and aligns with Industry 5.0's human-centric values \cite{DBLP:journals/access/QiT19}. Additionally, this setup reduces dependence on network connectivity compared with cloud-based processing, making it well-suited for latency-sensitive tasks \cite{DBLP:conf/ictc/SampedroMSA0L22}. This operating method also reduces network congestion, essential for wireless network performance \cite{DBLP:journals/access/DonnoTD19}.

However, implementing Edge AI is far from being trivial and presents several challenges \cite{singh2023edge}. For example, hardware must support AI workloads on devices with limited capacity, distributed resources need efficient management, and interoperability across varied edge environments must be guaranteed. Security measures are critical since edge devices often operate under less controlled conditions than centralized servers. Meeting these requirements calls for new toolkits that facilitate deployment, optimize resource use, and guarantee system reliability \cite{DBLP:journals/internet/MeuserLBPDABBLDEGMPRSSS24,DBLP:conf/icccn/VaterHK19}.

In response to some of these needs, we propose a novel framework for the fast deployment of Edge AI solutions in industrial scenarios. Therefore, the main contributions of this research are:

\begin{itemize}
	\item We present a solution for fast deployment of edge AI in Industry 5.0 applications. This solution brings several innovations around the concept of Collaborative Intelligence (CI), as it allows the joint work of an operator together with an intelligent agent to curate AI models deployed at the edge.

	\item We validate our approach through a use case within an industrial setting in the food industry domain. The validation includes observations on performance gains, compatibility with existing industrial infrastructure, and flexibility in adapting to various operational scenarios.

	\item We examine existing challenges and propose future research areas for Edge AI in industrial systems that prioritize human interaction. These areas cover developing efficient algorithms for hardware with limited resources and improving user interaction with intelligent industrial technologies.
\end{itemize}

The remainder of this paper proceeds as follows. Section 2 examines state-of-the-art cloud, edge, and fog computing alongside AI integration. Section 3 introduces our proposed framework for fast Edge AI deployment and explains how it addresses the gaps in current methods. Section 4 evaluates our proposed approach through a food industry case study. Section 5 compares this approach with existing ones and discusses challenges and opportunities for Edge AI in practical settings. Section 6 concludes with the lessons learned from this research and outlines future research directions.

\section{State-of-the-Art}
This section explores how existing approaches address current challenges, focusing on their ability to support real-time operations, improve accessibility, and meet security requirements. Additionally, we examine how integrating AI with existing computing models has opened new possibilities, mainly in applications that demand fast data processing. The role of human involvement in AI-driven processes is also considered a proper way to balance automation with expertise.

\subsection{Antecedents}
Cloud, fog, and edge computing development have altered how industries process and apply data \cite{DBLP:journals/iotj/HuJYLLWZXY24}. Cloud computing offers internet-based access to shared resources, securing remote availability, enabling automatic scaling to meet service agreements, and providing serverless billing for actual usage \cite{DBLP:journals/access/DonnoTD19,DBLP:journals/software/MakitaloOKAKM17}. Fog computing positions processing closer to gateways and routers, reducing delay, lowering data transfers to central servers, and improving responsiveness while saving network capacity \cite{DBLP:conf/icca3/Al-QamashSAS18,DBLP:journals/access/QiT19}. Edge computing processes data at its source in IoT and sensor networks, cutting latency, avoiding central bottlenecks, and offering simpler deployment with stronger security for real-time, localized tasks \cite{DBLP:journals/internet/MeuserLBPDABBLDEGMPRSSS24,DBLP:journals/cm/ShenSZLPLZL24}.

Combining AI and edge computing provides significant benefits \cite{shi2020communication}. Processing data at the edge enables real-time responsiveness essential for applications requiring minimal delay, including autonomous systems \cite{DBLP:journals/ijon/SunWMZ25}. Also, handling sensitive information locally improves privacy, making it suitable for biometric data or personal details applications \cite{gill2025edge}. This approach aligns with the goals of Industry 5.0, enabling the creation of more innovative applications while ensuring security and efficiency in industrial systems \cite{DBLP:journals/access/BuchgeherGGE21,DBLP:journals/frai/KrausePKKLESUDKDGHHGSM24}.

\subsection{The Present}
Edge AI is gaining even more attention as organizations transition toward Industry 5.0. Recent research has focused on applying large language models (LLMs) to edge devices, enabling an efficient resource allocation in connected environments \cite{DBLP:journals/cm/ShenSZLPLZL24}. Convolutional neural networks (CNNs) have also been tailored for edge deployment to address limitations in computational power and ensure efficient operation without compromising performance \cite{DBLP:journals/ijon/SunWMZ25}. 

Deploying Edge AI in industrial domains requires balancing the computational needs of algorithms with the constraints of devices operating on-site \cite{deng2020edge}. Dynamic neural network (DNN) partitioning strategies address these challenges by distributing workloads across edge and cloud environments, reducing energy demand, and improving scalability \cite{DBLP:journals/jiii/GauttamPBNP23}. Studies assessing the performance of diverse edge devices show the importance of establishing clear benchmarks and architectural guidelines \cite{DBLP:conf/case/RupprechtHV22}. The interplay between edge, cloud, and fog computing has shown the need for effective coordination among these layers to support seamless operations \cite{DBLP:journals/access/DonnoTD19}.

An additional focus is on integrating human expertise with AI-driven systems. In this regard, knowledge graphs are being explored to improve collaboration and provide alignment with the goals of Industry 5.0 \cite{martinez2024examining,DBLP:journals/frai/KrausePKKLESUDKDGHHGSM24}. This approach aims to create efficient systems that are adaptable to human needs and capable of facilitating trust. These advancements reveal the importance of sustainable designs prioritizing long-term effectiveness \cite{DBLP:journals/jeim/ChenLXPY22}.

\subsection{Open Challenges}
CI now involves scenarios where humans refine AI outputs, as seen in human-in-the-loop (HITL) systems \cite{DBLP:conf/hhai/HartikainenSV24}. Humans and AI collaborate in co-creation, with AI providing data insights. Augmented intelligence improves human abilities by supporting people with AI-driven real-time assistance. Existing works are focused on demonstrating the evolving synergy between humans and AI \cite{ammann2025automated}.

Although research continues to propose new solutions, several challenges persist. Incorporating human involvement increases costs and limits scalability, making implementation more complex. Guaranteeing that contributors have adequate training and expertise is another critical issue, as their input directly affects system outcomes. Lastly, achieving the right balance between automation and human input remains a significant task to ensure optimal performance. 

\subsection{Contribution Over the State-of-the-art}
Our proposed framework introduces a novel approach to rapidly deploying Edge AI solutions tailored for Industry 5.0 applications. It offers a modular design compatible with diverse edge devices, making it suitable for various manufacturing domains. Processing AI tasks directly at the edge allows for lower latency and strengthens data privacy by reducing reliance on public networks. It uses real-time protocols for performance monitoring, enabling the (automatic) correction of system irregularities in critical scenarios. Our pre-built modules optimized for edge environments address computational challenges while maintaining results. In addition, one of the new features is the agentic architecture that allows some tasks to be performed by an agent, whether human, computational or even the result of a collaboration.

\section{A Framework for Rapid Deployment of Edge AI Solutions}
Our framework enables organizations to rapidly prototype, test, and deploy Edge AI models across various industrial settings. It integrates best practices for Edge AI, allowing seamless communication between edge devices and centralized servers. The framework integrates multiple technologies to facilitate real-time data analysis. It uses real-time protocols (e.g., MQTT) to connect edge devices to a broker, handling a bidirectional data flow. It also incorporates interactive data visualization to monitor system performance through dynamic charts. 

The system supports quick prototyping and deployment of AI models directly on edge hardware. It combines established communication protocols with a modular design to facilitate local inference and immediate data feedback. Edge devices connect to an MQTT broker that manages message flow in both directions. The system also includes a web-based interface that displays prediction results and system performance through real-time visualizations.

The main features include real-time sensor data ingestion, a browser-based table displaying actual and predicted values with status flags (OK / Non-OK), and automatic interpretation of prediction errors using external AI services. The user operator can always use a web interface, enabling automatic and manual recalibration and the chance to inspect detected anomalies further.

\subsection{Architecture}
\begin{figure}
	\centering
		\includegraphics[width=1.00\textwidth]{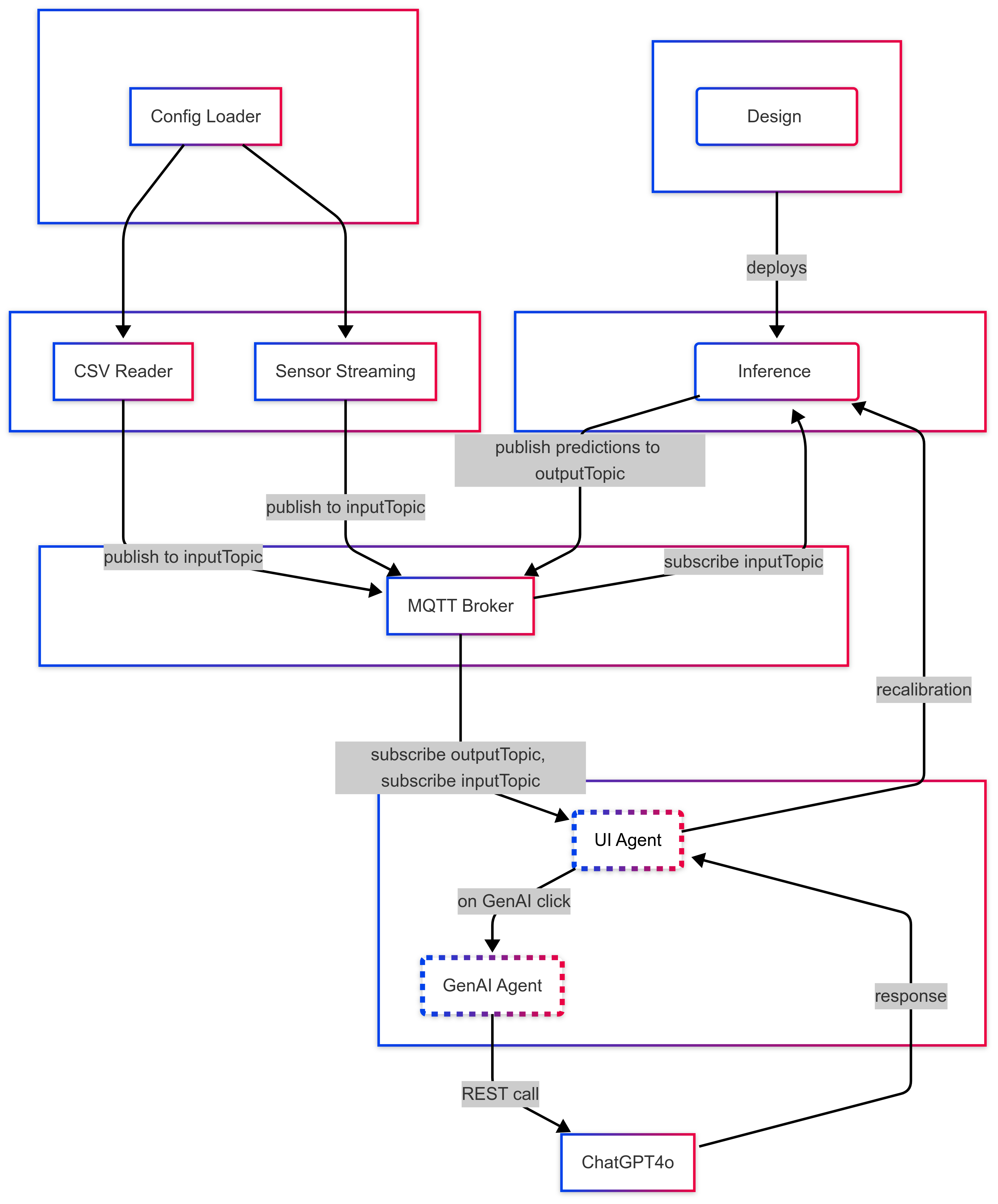}
	\caption{Our framework begins when the Config Loader initializes both CSV Reader and Sensor Streaming, routes data through the MQTT Broker to the Inference component, returns predictions to the UI Agent, invokes the GenAI Agent via ChatGPT4o for on-demand analysis, and uses the Design component to deploy updates to the inference component}
	\label{fig:arc}
\end{figure}

Figure \ref{fig:arc} shows our architecture, where each component handles a different task. Initially, the \textbf{Config Loader} sets up the configurations that feed into data ingestion sources: \textbf{CSV Reader} for static datasets and \textbf{Sensor Streaming} for real-time data. Both sources publish data to the \textbf{MQTT Broker} under the \texttt{inputTopic}. 

Within the \textbf{Processing} stage, an \textbf{Inference Agent} subscribes to the incoming data, processes it, and publishes predictions back to the \texttt{outputTopic}. The \textbf{UI Agent} in the \textbf{Presentation} stage subscribes to raw data and inference outputs for immediate visualization. Additionally, a \textbf{GenAI Agent} integrates with external generative models via REST API calls (e.g., ChatGPT4o) to provide improved user interactions upon request.

A dedicated \textbf{Designer Agent} within the \textbf{Pipeline} stage manages the deployment of inference pipelines. The \textbf{UI Agent} further allows recalibration commands directly affecting the \textbf{Inference Agent}. Components with dashed outlines represent interactive or adaptable elements, with a strong focus on user-driven configuration.

This setup assigns one responsibility per component, so adding new data sources or visualizations does not require changes elsewhere. The automated deployment pipeline ensures that updates to the inference logic are tested and rolled out without manual intervention. On-demand AI explanations appear seamlessly in the UI, helping users understand model outputs. The design enables straightforward maintenance and gradual improvements.

\subsection{Initialization}
The framework communicates with edge devices using the MQTT protocol and responds to incoming data streams in real-time. A configuration file, provided in JSON format, specifies essential parameters, including the broker address, topic names, and the structure of the input features. Once this configuration is loaded, the application automatically connects to the broker, subscribes to the defined input and output topics, and begins data exchange. The messaging component handles the connection lifecycle and facilitates the publishing of sensor data as well as the reception of prediction results. Topics are created dynamically based on the configuration, and the interface provides feedback on the current connection status and broker interactions.

\begin{figure}
	\centering
		\includegraphics[width=0.98\textwidth]{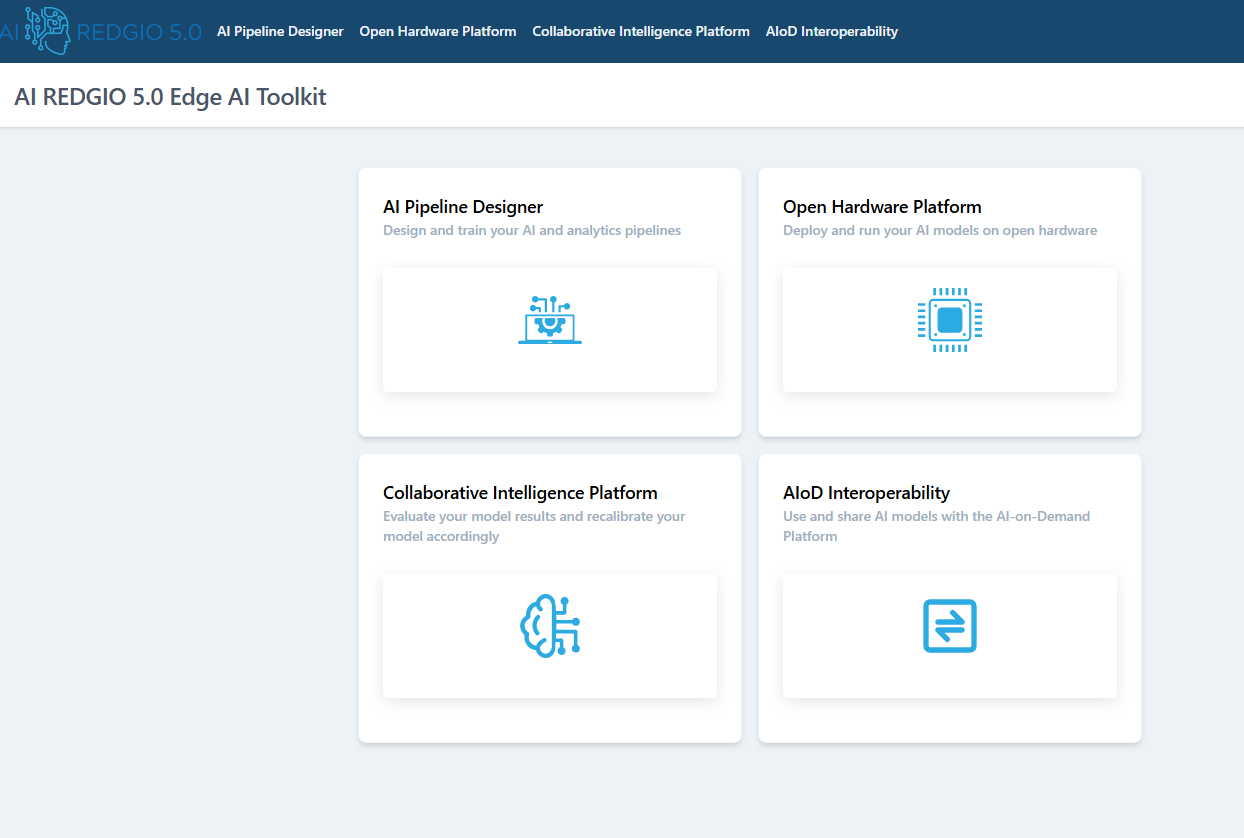}
	\caption{Start screen available to the operator to initialize each of the components of the framework}
	\label{fig:tk}
\end{figure}

Figure \ref{fig:tk} shows the screen with which operators can initialize the various components of the framework, including the CI component (with the loading module), the Open Hardware (device(s) running on the edge), the Design component to design and transfer the AI models to the device(s) with ease. Moreover, extra documentation about models that could be used.

\subsection{UI Agent}
The UI Agent provides a real-time web interface that displays incoming data streams and model outputs in a clear, interactive format. At the core of the interface is a table that presents each data row alongside the model's predictions. The target values in this table are editable, enabling users to correct or adjust them manually when necessary. This feature supports HITL workflows, simplifying results based on domain knowledge or observed discrepancies.

Beyond the table, the interface includes two visual components: a time series chart and a bar chart. The time series chart tracks predicted versus actual values, offering a temporal view of model accuracy. The bar chart categorizes entries as OK or Non-OK, providing a quick overview of current classification trends. Both visualizations update automatically with each new data point, allowing users to monitor system performance and detect real-time anomalies.

\begin{figure}
	\centering
		\includegraphics[width=1.01\textwidth]{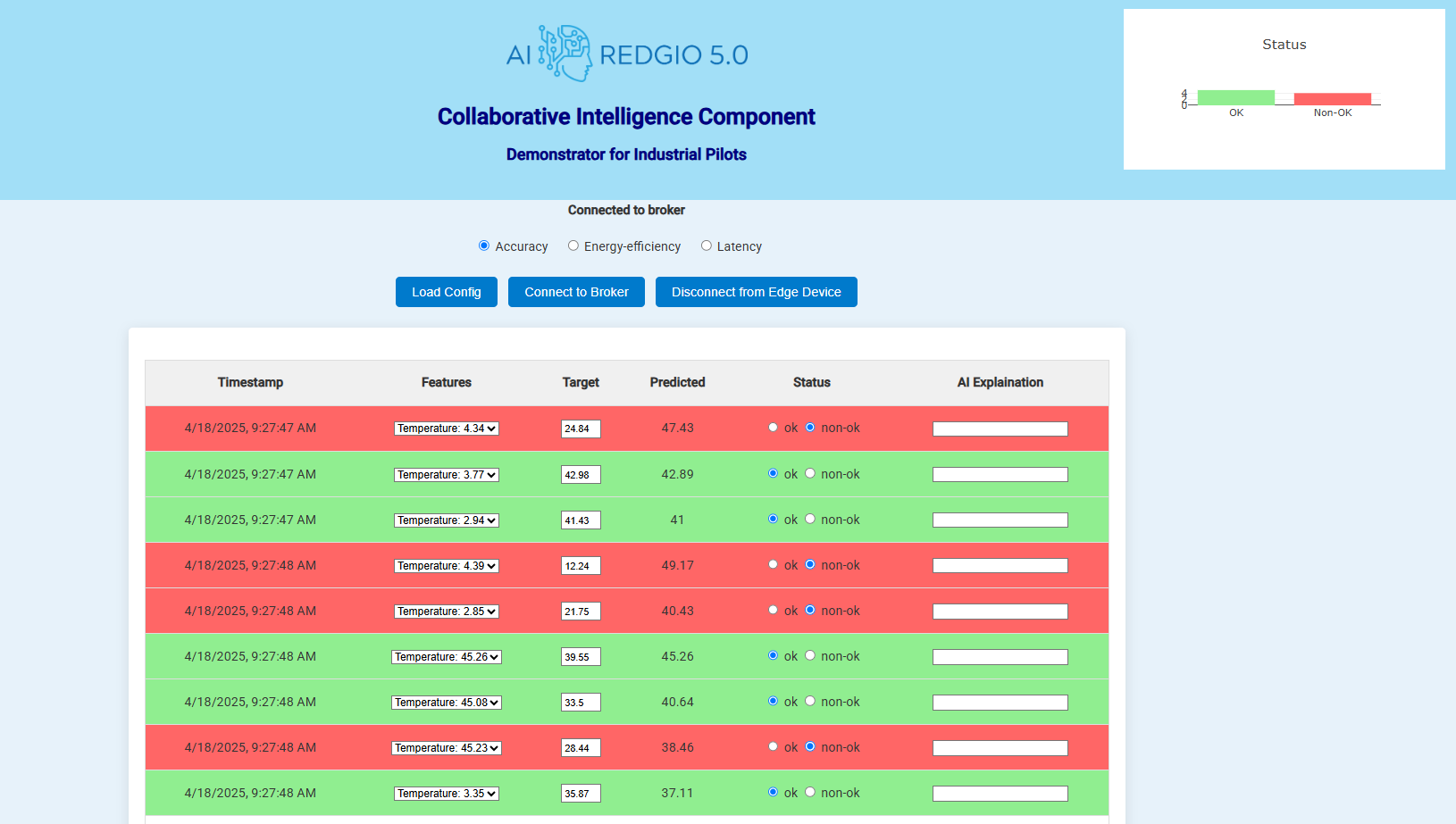}
	\caption{Browser-based screenshot of the CI component with the functionality that allows monitoring and correction of the models hosted on the Edge}
	\label{fig:ci}
\end{figure}

Figure \ref{fig:ci} shows the interface, enabling automatic and manual recalibration and the possibility of inspecting detected anomalies further. In addition, if a large training data set is unavailable for the model at the edge, the operator can artificially generate and curate training data.

\begin{figure}
	\centering
		\includegraphics[width=1.01\textwidth]{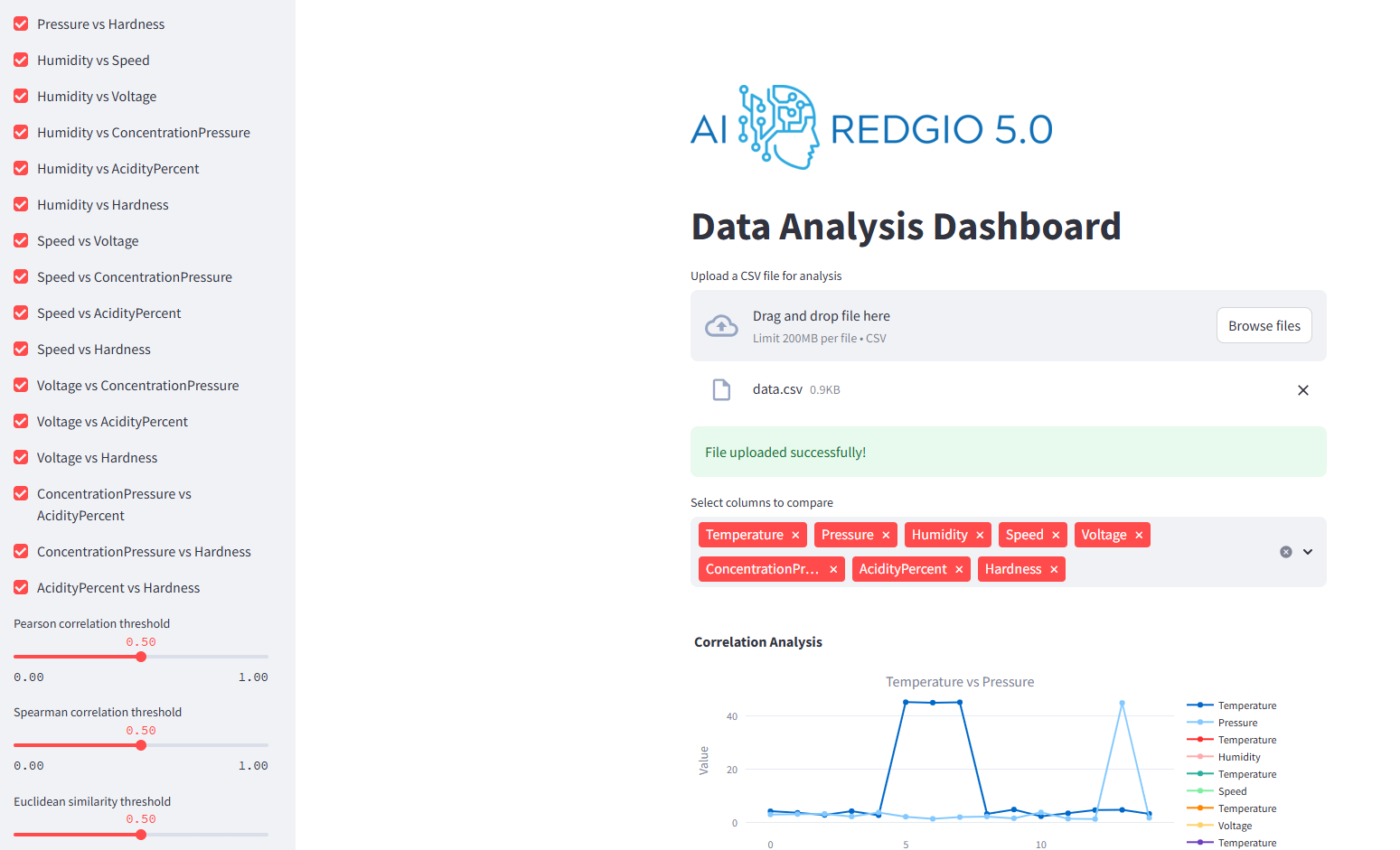}
	\caption{Data analysis features of the CI component, showing updated graphs next to controls in a split-view layout}
	\label{fig:das}
\end{figure}

Figure \ref{fig:das} shows another aspect of the UI agent that allows the visualization of the data loaded. Human operators can interact with dropdown menus and sliders to filter data based on the attributes of their datasets, updating visualizations automatically. The interface includes a sidebar with controls and a central display area featuring knowledge graphs. This structure supports interactive exploration, querying, and basic reasoning over the dataset.

\subsection{GenAI Agent}
When triggered, the GenAI Agent constructs a structured prompt containing a subset of the most relevant features, including the raw input vector, the model's predicted output, the expected target value, and a confidence score, if available. These values are serialized into a JSON payload and transmitted to an external large language model (e.g., GPT-4o) via a RESTful API endpoint. The request includes metadata fields that define the task type (e.g., explain prediction, assist with labeling), the expected format of the response (plain text, structured tags, etc.), and any domain-specific instructions provided by the operator.

The GenAI agent monitors prediction outputs for discrepancies beyond a configurable error threshold (e.g., absolute or percentage deviation). The GenAI Agent automatically dispatches the relevant context to the AI backend when such a deviation is detected. Upon receiving the response, the explanation is parsed and embedded in the corresponding entry within the data stream. This explanation is usually a concise natural language justification of the model’s decision and is intended to be easily understood by human operators. The entry is simultaneously flagged for potential recalibration, and a visual indicator is rendered in the user interface.

The component maintains a rolling buffer of unlabeled instances or weakly labeled entries in AI-assisted labeling mode. It constructs batch prompts that request class assignments based on predefined criteria. The GenAI Agent supports zero-shot and few-shot prompting strategies, where prior labeled examples can be included to improve consistency. The prompt templates for explanation are stored in external configuration files and can be modified at runtime. This enables domain experts to adapt the system’s language, constraints, and verbosity without altering the codebase. 

\subsection{Inference Component}
The Inference Component runs on an edge device (it was successfully tested on ESP32 and Raspberry Pi when preparing this work) for low-power, wireless operation in edge environments. Its open hardware foundation allows developers to modify and extend the device to meet specific use cases, making it suitable for various industrial settings.

This agent applies trained models to incoming data and returns predictions. It communicates through MQTT for lightweight messaging and supports HTTP-based APIs over Wi-Fi for additional data exchange needs. Built-in libraries simplify integration and protocol handling, allowing the device to function reliably as both a data receiver and a publisher in real-time processing workflows.

\subsection{Design Component}
The Design Component acts as a pipeline builder being able to support developing and deploying AI pipelines for execution on edge or cloud infrastructure. It is accessible to users with varying levels of technical experience. Users can perform basic data preparation, run models for industrial analytics, or generate visual outputs. The component allows exporting results in formats compatible with other systems, such as downloadable files or API responses.

The component includes configurable modules for data processing and model execution. Users can select specific operations, modify parameters, and assemble them into complete pipelines. These can be scheduled or triggered by events with built-in validation steps. Once trained, models can be deployed to edge devices for on-site inference. The platform tracks performance and outcome metrics during execution, offering continuous feedback. External models built with supported libraries can also be integrated and deployed within the same environment, allowing flexibility in adapting to different requirements.

\subsection{Components Working Together}
The framework establishes a connection to an MQTT broker chosen for its efficiency in handling low-latency and low-bandwidth communication. Once connected, it subscribes to a predefined topic and receives messages from edge devices or sensors. These messages typically contain structured observations such as temperature values or other sensor readings. As data arrives, the framework processes and visualizes it in real time. Internal handlers manage events related to the connection state, ensuring resilience during operation.

Incoming data is rendered through two interactive charts. The first is a time-series plot that tracks predicted and actual values over time, helping users identify deviations and monitor the performance of deployed models. The second is a status chart as a bar plot, summarizing the number of data points classified as OK or Non-OK. These visualizations provide an overview of system behavior and help detect anomalies or drift in model predictions.

In addition to receiving data, the system can publish sensor readings or prediction results to the broker for downstream processing. All incoming messages are validated using predefined schemas before they are displayed. The platform also supports ingesting historical data through CSV file uploads, which are streamed to the broker as if generated in real-time. Data points should be labeled based on expert criteria for integrity monitoring and flagged as OK or Non-OK. These classifications appear in both the table view and the visual summaries, and users can export all Non-OK entries as JSON for further analysis or documentation.

\section{Use Case}
Gradiant, a technological center in Galicia, is supporting Quescrem, a leading company in the production of cream cheese, in the adoption of this framework to improve monitoring and control within their cheese production processes. The framework gathers real-time data at different production stages, including milk processing, homogenization, and packaging. Sensors monitor variables such as fat content, pH levels, pressure, and temperature, transmitting the information through specified input topics. The human operator and the AI-based agent analyze this data, predicting potential deviations and classifying batches as OK or Non-OK. Operators can adjust targets and record explanations for Non-OK batches, maintaining transparent decision-making.

This approach is especially beneficial for the company in maintaining consistent cream cheese quality, detecting irregularities early, and preventing large-scale production disruptions. Gradiant supplies the technological base, continuously refining AI predictions. Automated explanations clarify batch irregularities, aiding technicians in implementing corrective measures quickly. Structured JSON exports of Non-OK cases support further analysis, improving processes continuously and minimizing waste. This method results in an efficient, data-driven workflow consistent with high-quality product standards. Automated deployment accelerates system setup and reduces latency, which is essential for real-time manufacturing operations.

Deploying this framework allows the cheese-making company to improve production effectiveness substantially. Analyzing localized sensor data enables proactive maintenance, decreasing downtime by approximately 65\% due to timely issue detection. Optimized machine utilization yields around 20\% better energy efficiency, promoting more sustainable production practices.

Performance evaluations employed controlled industrial simulations using sensor-generated data streams, concentrating on essential performance metrics. Deployment setup time decreased by around 80\% compared to traditional manual procedures. Additionally, the framework maintained average end-to-end latencies under 200 milliseconds, supporting rapid data processing. Predictive accuracy consistently exceeded 95\%, confirming reliability.

\section{Comparison with Related Systems}
We compare our system with \textbf{AIfES}~\cite{wulfert2024aifes}, \textbf{InfiniEdge AI}\footnote{\url{https://github.com/lfedgeai}}, and \textbf{OpenEI}~\cite{zhang2019openei}. The reason is that these frameworks represent well-known approaches for deploying ML models on embedded or resource-constrained devices. AIfES focuses on deploying neural networks on ultra-low-power microcontrollers and offers native C++ implementations suitable for embedded platforms. It performs well under tight resource constraints but does not include built-in support for networked communication or visualization tools. InfiniEdge AI is an open-source system designed for edge computing environments with more capacity. It allows model deployment across different nodes and supports container-based deployment pipelines. Its focus is on scalability and compatibility with existing edge infrastructure. OpenEI offers a lightweight setup for distributed inference but limits its operation to simpler data flows and lacks modular extension points for integrating additional services or tools. Some comparative differences are:

\begin{itemize}
	\item Regarding latency, our solution preserves low latency via MQTT messaging. AIfES does not define specific latency targets. InfiniEdge AI supports real-time performance. OpenEI also handles low latency.

	\item Regarding modularity, our approach uses plug-and-play components to cover various uses. AIfES and OpenEI address small setups. InfiniEdge AI scales in open-source environments but offers fewer interchangeable parts.

	\item Regarding data handling and privacy, our framework processes data locally, keeping it off external networks. AIfES also runs locally but with limited features. InfiniEdge AI includes strong security measures, being the strictest in this aspect. OpenEI offers only minimal privacy safeguards.

	\item Regarding target environments, our system fits human-centric industrial sites requiring real-time decision support. AIfES focuses strongly on ultra-low-resource devices. InfiniEdge AI covers multiple industries. OpenEI works best in lab or demo settings.

	\item Regarding visualization, we offer live charts and performance monitors. AIfES provides little to no visualization. InfiniEdge AI has moderate built-in tools. OpenEI handles small datasets with simple displays.

	\item Regarding cost and hardware requirements, our approach runs on modest edge hardware and pairs with open-source tools to keep costs low. AIfES and OpenEI also aim to design at a low cost. InfiniEdge AI is also open-source but needs more setup.
\end{itemize}

In addition to these differences, our framework is the first of its kind to implement an agent-based strategy, which means that some specialized tasks can be performed by the human operator, the machine or both in collaboration, which, together with the other features, makes it fit the values of Industry 5.0.

\section{Conclusions and Future Work}

This paper has presented a framework for Edge AI suited to Industry 5.0. It supports real-time, local data handling, which helps reduce delays, lower network usage, and improve data protection. Our proposed approach enables low-latency processing and improved data privacy by performing inference locally, eliminating dependence on remote servers. It also enables real-time human interaction and supports manual and automatic recalibration at the edge.

Our framework allows a broad range of industrial domains to integrate new functionality without significant infrastructure changes. The agent-based design further simplifies deployment, enabling rapid configuration and extension of capabilities as operational needs evolve.

However, we have also seen that deploying our framework comes with significant hurdles. Edge devices often have limited computational and storage capabilities, requiring efficient AI models that preserve accuracy. Connecting Edge AI to existing systems can be difficult, as many setups lack support. Data from edge devices may be incomplete or unreliable, which affects performance. Upfront hardware, software, and training costs remain a barrier, especially for smaller organizations.

Future work will improve model efficiency through techniques such as quantization, extend deployment to more heterogeneous edge environments, and refine the GenAI agent’s interaction through feedback-driven prompt adaptation. We also plan to validate the framework in additional industrial domains to assess generalizability and scalability.

\section*{Supplementary Material}
The source code for the CI component is available under an open-source license at \url{https://github.com/AI-REDGIO-5-0/ci-component}. The repository includes documentation, installation instructions, and example configurations to support reuse and contributions. A user reference video is available at \url{https://www.youtube.com/watch?v=AR8F8U-QXhM}, and additional information about the assets used in this research (Designer component, hardware, etc.) can be found at \url{https://wiki.ai-redgio50.s5labs.eu}.

\section*{Acknowledgements}
We would like to thank in advance the anonymous reviewers for their help in improving the manuscript. This work has been supported by the European Union’s Horizon Europe research and innovation programme under grant agreement No [101092069](Regions and (E)DIHs alliance for AI-at-the-Edge adoption by European Industry 5.0 Manufacturing SMEs [AI REDGIO 5.0]).

\bibliographystyle{plain}
\bibliography{mybib}

\begin{thebibliography}{10}

\bibitem{DBLP:conf/icca3/Al-QamashSAS18}
Amal Al{-}Qamash, Iten Soliman, Rawan Abulibdeh, and Moutaz Saleh~Mustafa
  Saleh.
\newblock Cloud, fog, and edge computing: {A} software engineering perspective.
\newblock In {\em 2018 International Conference on Computer and Applications
  (ICCA), Beirut, Lebanon, August 25-26, 2018}, pages 276--284. {IEEE}, 2018.

\bibitem{ammann2025automated}
Lolita Ammann, Jorge Martinez-Gil, Michael Mayr, and Georgios~C Chasparis.
\newblock Automated knowledge graph learning in industrial processes.
\newblock {\em Procedia Computer Science}, 253:2428--2437, 2025.

\bibitem{DBLP:journals/access/BuchgeherGGE21}
Georg Buchgeher, David Gabauer, Jorge~Mart{\'{\i}}nez Gil, and Lisa Ehrlinger.
\newblock Knowledge graphs in manufacturing and production: {A} systematic
  literature review.
\newblock {\em {IEEE} Access}, 9:55537--55554, 2021.

\bibitem{DBLP:journals/jeim/ChenLXPY22}
Wenting Chen, Caihua Liu, Fei Xing, Guochao Peng, and Xi~Yang.
\newblock Establishment of a maturity model to assess the development of
  industrial {AI} in smart manufacturing.
\newblock {\em J. Enterp. Inf. Manag.}, 35(3):701--728, 2022.

\bibitem{DBLP:conf/interact/BerghRG23}
Jan~Van den Bergh, Jorge Rodr{\'{\i}}guez{-}Echeverr{\'{\i}}a, and Sidharta
  Gautama.
\newblock Towards a smart combination of human and artificial intelligence for
  manufacturing.
\newblock In Anna Bramwell{-}Dicks, Abigail Evans, Marco Winckler, Helen
  Petrie, and Jos{\'{e}}~L. Abdelnour{-}Nocera, editors, {\em Design for
  Equality and Justice - {INTERACT} 2023 {IFIP} {TC} 13 Workshops, York, UK,
  August 28 - September 1, 2023, Revised Selected Papers, Part {I}}, volume
  14535 of {\em LNCS}, pages 20--30. Springer, 2023.

\bibitem{deng2020edge}
Shuiguang Deng, Hailiang Zhao, Weijia Fang, Jianwei Yin, Schahram Dustdar, and
  Albert~Y Zomaya.
\newblock Edge intelligence: The confluence of edge computing and artificial
  intelligence.
\newblock {\em IEEE Internet of Things Journal}, 7(8):7457--7469, 2020.

\bibitem{DBLP:journals/access/DonnoTD19}
Michele~De Donno, Koen Tange, and Nicola Dragoni.
\newblock Foundations and evolution of modern computing paradigms: Cloud, iot,
  edge, and fog.
\newblock {\em {IEEE} Access}, 7:150936--150948, 2019.

\bibitem{DBLP:journals/jiii/GauttamPBNP23}
Himanshu Gauttam, Kiran~Kumar Pattanaik, Saumya Bhadauria, Garima Nain, and
  Putta~Bhanu Prakash.
\newblock An efficient {DNN} splitting scheme for edge-ai enabled smart
  manufacturing.
\newblock {\em J. Ind. Inf. Integr.}, 34:100481, 2023.

\bibitem{gill2025edge}
Sukhpal~Singh Gill, Muhammed Golec, Jianmin Hu, Minxian Xu, Junhui Du, Huaming
  Wu, Guneet~Kaur Walia, Subramaniam~Subramanian Murugesan, Babar Ali, Mohit
  Kumar, et~al.
\newblock Edge ai: A taxonomy, systematic review and future directions.
\newblock {\em Cluster Computing}, 28(1):1--53, 2025.

\bibitem{DBLP:conf/hhai/HartikainenSV24}
Maria Hartikainen, Guna Spurava, and Kaisa V{\"a}{\"a}n{\"a}nen.
\newblock Human-ai collaboration in smart manufacturing: Key concepts and
  framework for design.
\newblock pages 162--172. IOS Press, 2024.

\bibitem{hoch2023multi}
Thomas Hoch, Jorge Martinez-Gil, Mario Pichler, Agastya Silvina, Bernhard
  Heinzl, Bernhard Moser, Dimitris Eleftheriou, Hector~Diego Estrada-Lugo, and
  Maria~Chiara Leva.
\newblock Multi-stakeholder perspective on human-ai collaboration in industry
  5.0.
\newblock In {\em Artificial Intelligence in Manufacturing: Enabling
  Intelligent, Flexible and Cost-Effective Production Through AI}, pages
  407--421. Springer Nature Switzerland Cham, 2023.

\bibitem{DBLP:journals/iotj/HuJYLLWZXY24}
Yujiao Hu, Qingmin Jia, Yuan Yao, Yong Lee, Mengjie Lee, Chenyi Wang, Xiaomao
  Zhou, Renchao Xie, and F.~Richard Yu.
\newblock Industrial internet of things intelligence empowering smart
  manufacturing: {A} literature review.
\newblock {\em {IEEE} Internet Things J.}, 11(11):19143--19167, 2024.

\bibitem{DBLP:journals/frai/KrausePKKLESUDKDGHHGSM24}
Franz Krause, Heiko Paulheim, Elmar Kiesling, Kabul Kurniawan, Maria~Chiara
  Leva, Hector~Diego Estrada{-}Lugo, Gernot St{\"{u}}bl, Nazim~Kemal {\"{U}}re,
  Javier Dominguez{-}Ledo, Maqbool Khan, Pedro Demolder, Hans Gaux, Bernhard
  Heinzl, Thomas Hoch, Jorge~Mart{\'{\i}}nez Gil, Agastya Silvina, and
  Bernhard~A. Moser.
\newblock Managing human-ai collaborations within industry 5.0 scenarios via
  knowledge graphs: key challenges and lessons learned.
\newblock {\em Frontiers Artif. Intell.}, 7, 2024.

\bibitem{DBLP:journals/software/MakitaloOKAKM17}
Niko M{\"{a}}kitalo, Aleksandr Ometov, Joona Kannisto, Sergey Andreev, Yevgeni
  Koucheryavy, and Tommi Mikkonen.
\newblock Safe, secure executions at the network edge: Coordinating cloud,
  edge, and fog computing.
\newblock {\em {IEEE} Softw.}, 35(1):30--37, 2018.

\bibitem{martinez2024examining}
Jorge Martinez-Gil, Thomas Hoch, Mario Pichler, Bernhard Heinzl, Bernhard
  Moser, Kabul Kurniawan, Elmar Kiesling, and Franz Krause.
\newblock Examining the adoption of knowledge graphs in the manufacturing
  industry: A comprehensive review.
\newblock {\em Artificial Intelligence in Manufacturing}, page~55, 2024.

\bibitem{DBLP:journals/internet/MeuserLBPDABBLDEGMPRSSS24}
Tobias Meuser, Lauri Lov{\'{e}}n, Monowar~H. Bhuyan, Shishir~G. Patil, Schahram
  Dustdar, Atakan Aral, Suzan Bayhan, Christian Becker, Eyal de~Lara, Aaron~Yi
  Ding, Janick Edinger, James Gross, Nitinder Mohan, Andy~D. Pimentel, Etienne
  Rivi{\`{e}}re, Henning Schulzrinne, Pieter Simoens, G{\"{u}}rkan Solmaz, and
  Michael Welzl.
\newblock Revisiting edge {AI:} opportunities and challenges.
\newblock {\em {IEEE} Internet Comput.}, 28(4):49--59, 2024.

\bibitem{DBLP:journals/access/QiT19}
Qinglin Qi and Fei Tao.
\newblock A smart manufacturing service system based on edge computing, fog
  computing, and cloud computing.
\newblock {\em {IEEE} Access}, 7:86769--86777, 2019.

\bibitem{DBLP:conf/case/RupprechtHV22}
Bernhard Rupprecht, Dominik Hujo, and Birgit Vogel{-}Heuser.
\newblock Performance evaluation of {AI} algorithms on heterogeneous edge
  devices for manufacturing.
\newblock In {\em 18th {IEEE} International Conference on Automation Science
  and Engineering, {CASE} 2022, Mexico City, Mexico, August 20-24, 2022}, pages
  2132--2139. {IEEE}, 2022.

\bibitem{DBLP:conf/ictc/SampedroMSA0L22}
Gabriel Avelino~R. Sampedro, Raymer Manaig, Paulyne Salazar{-}Manaig, Mideth~B.
  Abisado, Dong{-}Seong Kim, and Jae{-}Min Lee.
\newblock An overview of opportunities and challenges of edge computing in
  smart manufacturing.
\newblock In {\em 13th International Conference on Information and
  Communication Technology Convergence, {ICTC} 2022, Jeju Island, Korea,
  Republic of, October 19-21, 2022}, pages 2182--2186. {IEEE}, 2022.

\bibitem{DBLP:journals/cm/ShenSZLPLZL24}
Yifei Shen, Jiawei Shao, Xinjie Zhang, Zehong Lin, Hao Pan, Dongsheng Li, Jun
  Zhang, and Khaled~B. Letaief.
\newblock Large language models empowered autonomous edge {AI} for connected
  intelligence.
\newblock {\em {IEEE} Commun. Mag.}, 62(10):140--146, 2024.

\bibitem{shi2020communication}
Yuanming Shi, Kai Yang, Tao Jiang, Jun Zhang, and Khaled~B Letaief.
\newblock Communication-efficient edge ai: Algorithms and systems.
\newblock {\em IEEE Communications Surveys \& Tutorials}, 22(4):2167--2191,
  2020.

\bibitem{singh2023edge}
Raghubir Singh and Sukhpal~Singh Gill.
\newblock Edge ai: a survey.
\newblock {\em Internet of Things and Cyber-Physical Systems}, 3:71--92, 2023.

\bibitem{DBLP:journals/ijon/SunWMZ25}
Kailai Sun, Xinwei Wang, Xi~Miao, and Qianchuan Zhao.
\newblock A review of {AI} edge devices and lightweight {CNN} and {LLM}
  deployment.
\newblock {\em Neurocomputing}, 614:128791, 2025.

\bibitem{DBLP:conf/icccn/VaterHK19}
Johannes Vater, Lars Harscheidt, and Alois~C. Knoll.
\newblock A reference architecture based on edge and cloud computing for smart
  manufacturing.
\newblock In {\em 28th International Conference on Computer Communication and
  Networks, {ICCCN} 2019, Valencia, Spain, July 29 - August 1, 2019}, pages
  1--7. {IEEE}, 2019.

\bibitem{wulfert2024aifes}
Lars Wulfert, Johannes K{\"u}hnel, Lukas Krupp, Justus Viga, Christian Wiede,
  Pierre Gembaczka, and Anton Grabmaier.
\newblock Aifes: A next-generation edge ai framework.
\newblock {\em IEEE Transactions on Pattern Analysis and Machine Intelligence},
  46(6):4519--4533, 2024.

\bibitem{zhang2019openei}
Xingzhou Zhang, Yifan Wang, Sidi Lu, Liangkai Liu, Weisong Shi, et~al.
\newblock Openei: An open framework for edge intelligence.
\newblock In {\em 2019 IEEE 39th International Conference on Distributed
  Computing Systems (ICDCS)}, pages 1840--1851. IEEE, 2019.

\end{thebibliography}

\end{document}